\documentclass{article} 
\usepackage{aditi}

\usepackage{microtype}
\usepackage{hyperref}
\usepackage{url}
\usepackage{booktabs}

\usepackage{style} 
\definecolor{skyblue}{RGB}{204,229,255}

\usepackage[most]{tcolorbox}
\tcbset{
  lightbluebox/.style={
    colback=skyblue!70,        
    colframe=skyblue!75!black, 
    boxrule=1pt,
    arc=4pt,
    auto outer arc
  }
}

\definecolor{darkblue}{rgb}{0, 0, 0.5}
\hypersetup{colorlinks=true, citecolor=darkblue, linkcolor=darkblue, urlcolor=darkblue}

\usepackage{titlesec}
\usepackage{url}
\usepackage{enumitem}
\usepackage{booktabs}       
\usepackage{amsfonts}       
\usepackage{nicefrac}       
\usepackage{microtype}      
\usepackage{subcaption}
\usepackage{xcolor}         
\usepackage{bm}             
\usepackage{graphicx}       
\usepackage{amsmath}        
\usepackage{amssymb}        
\usepackage{amsthm}         
\usepackage{xspace}         
\usepackage{adjustbox}      
\usepackage{multirow}       
\usepackage{pifont}         
\usepackage{caption}        
\usepackage{inconsolata}                                    
\usepackage{wrapfig}        
\usepackage{tcolorbox}
\usepackage{enumitem}
\usepackage[ruled,vlined]{algorithm2e}  
\usepackage{eqparbox}                                       
\newcommand{\countdown}{\texttt{Countdown}\xspace}             
\newcommand{\passat}[1]{Pass@$#1$\xspace}             

\renewcommand{\cite}[1]{\citep{#1}}                         

\titlespacing*{\subsection}
  {0pt}   
  {0.5\baselineskip}   
  {0.25\baselineskip} 

\titlespacing*{\section}
  {0pt}   
  {0.5\baselineskip}   
  {0.25\baselineskip} 

\title{Mode-Conditioning Unlocks Superior Test-Time Scaling}

\setauthors{Chen Henry Wu, \authorsep Sachin Goyal, \authorsep Aditi Raghunathan}
\setaffils{Carnegie Mellon University}

\setemail{\{chenwu2,sachingo,aditirag\}@cs.cmu.edu}
\setcode{https://github.com/AR-FORUM/modc}
\setwebsite{https://ar-forum.github.io/mod-c/}

\begin{document}

\maketitle

\begin{abstract}
Parallel sampling promises substantial gains in test-time scaling, but its effectiveness is sharply limited by diversity collapse, where models concentrate on a few modes and repeated samples produce the same mistakes. We propose the \textit{mode-conditioning (ModC) framework}, which explicitly allocates test-time compute across reasoning modes using either specialist models or mode-specific prefixes. ModC consistently improves scaling across controlled graph-search tasks and large-scale reasoning benchmarks, spanning model families and sizes from 0.5B to 7B. On OpenThoughts, fine-tuning Qwen2.5-7B with ModC achieves an \textit{4× efficiency gain} over standard training while also improving the maximum attainable \passat{k}. We further show that gradient clustering enables ModC without explicit mode labels, yielding up to $10\%$ gains on datasets such as NuminaMath. Finally, we show that ModC improves reinforcement learning (RL) and can further boost diversity-inducing RL methods. These results demonstrate that standard training underutilizes the diversity in data, and that ModC provides a simple, effective remedy for unlocking the full benefits of diversity in test-time scaling.
\end{abstract}

\section{Introduction}

Scaling test-time compute has become central to frontier reasoning systems, driving major advances in capability~\citep{wang2023selfconsistencyimproveschainthought,Snell2024ScalingLT,DeepSeekAI2025DeepSeekR1IR}. Test-time scaling can proceed either by lengthening individual reasoning traces or by \emph{parallel sampling}, where the model is given multiple independent attempts. Parallel scaling has proven especially effective~\citep{ma2025reasoningmodelseffectivethinking,brown2024largelanguagemonkeysscaling,wang2023selfconsistencyimproveschainthought}, and is particularly natural in domains like mathematics, coding, and scientific discovery, where candidate solutions can be verified automatically, making it a backbone of systems such as \citet{alphaevolve}.

Despite its promise, parallel scaling relies on a crucial assumption: the model must generate diverse and creative solutions. In practice, however, finetuning~\citep{dang2025weightensemblingimprovesreasoning} and reinforcement learning~\citep{yue2025doesreinforcementlearningreally} are well-documented to induce \textit{diversity collapse}, where generations concentrate on only a few dominant modes. As a result, additional samples often reproduce the same errors or converge on indistinguishable strategies, leading to diminishing returns as compute is scaled. While recent works propose ways to mitigate collapse, some important modes will almost inevitably remain underrepresented or assigned vanishing probability.

In this work, we put forward \textbf{mode-conditioning (ModC)}, a new paradigm that explicitly structures test-time scaling around multiple reasoning modes. Rather than drawing repeatedly from a collapsed distribution, we enforce coverage across strategies by conditioning on modes and allocating samples to cover diverse modes. This simple yet powerful -- and, to the best of our knowledge, previously unexplored -- idea provides a principled way to boost test-time scaling. Even when an LLM is balanced across two modes with complementary strengths, allocating $k/2$ samples to each mode strictly outperforms drawing $k$ samples from the original mixture. 
The advantage becomes especially pronounced on inputs where the dominant mode fails but a lower-probability one succeeds.

\begin{figure}[t!]
\centering
\begin{subfigure}{0.69\linewidth}
    \centering
    \includegraphics[width=\linewidth]{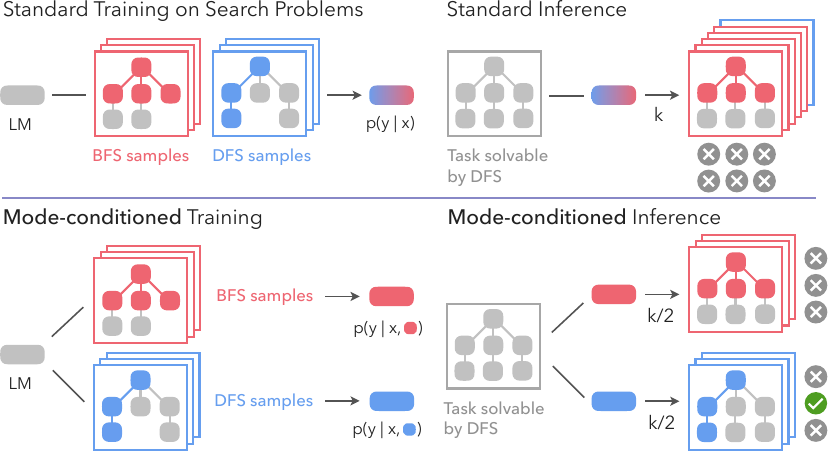}
\end{subfigure}
\hfill
\begin{subfigure}{0.28\linewidth}
    \centering
    \includegraphics[width=\linewidth]{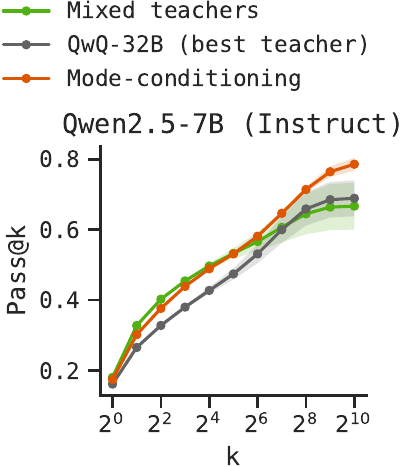}
\end{subfigure}
\caption{\label{fig:teaser} \textbf{Mode-conditioning for test-time scaling.} Modern LLMs often collapse to a single strategy, making \passat{k} scaling suboptimal: if the chosen strategy is wrong, every attempt fails. (Left) In a controlled graph task solvable by DFS or BFS, models trained on both still often commit to just one. To address this, we introduce \emph{mode-conditioning} (ModC) that explicitly allocates test-time compute across modes. We study two training methods that enable this: separate models or a single model with mode-specific prefixes. (Right) \textbf{4$\times$ efficiency gains with ModC training.} 
We apply ModC to long chain-of-thought reasoning distillation on the OpenThoughts dataset. With ModC, the model achieves the same Pass@1024 as standard training using only $k=256$ samples, 
yielding an $\sim$4$\times$ improvement in inference efficiency. Moreover, ModC also \emph{improves} the maximum attainable Pass@$k$, pushing the frontier of test-time scaling.
}
\end{figure}

While the principle of ModC is appealing, the key question is how to implement it in practice. One option is to prompt the model with explicit instructions for which mode to use, but this requires extensive manual effort to characterize modes and, more critically, models often fail to follow instructions for low-probability strategies. To move beyond such ad-hoc prompting, we train models to provide explicit control over modes. We explore two natural instantiations: (i) training \textit{separate specialist models}, each dedicated to a distinct mode, and (ii) training a single model with \textit{mode-specific prefixes}, where modes can be sampled reliably by using the corresponding prefix. These two versions offer different tradeoffs. The specialist approach ensures strong separation but eliminates opportunities for knowledge sharing across modes. By contrast, the prefix approach is more lightweight and enables positive transfer, but it can face capacity limits that prevent all modes from being fully captured, as well as imperfect control where the model fails to cleanly separate behaviors. Although both approaches are conceptually simple and have appeared in other contexts, we find that even these straightforward implementations yield substantial gains for test-time scaling.

We first consider a simple well-defined task of \countdown from~\citet{Gandhi2024StreamOS} which involves a search problem that admits two clear modes: breadth-first search (BFS) and depth-first search (DFS). In this well-defined controlled task, we can detect which mode a sample comes from by analyzing the trajectory. We see that standard training does in fact struggle to sample from both modes for several inputs. We also observe that dataset curation and model scaling offer only minimal gains. ModC with both separate models and prefixes shows \textbf{consistent superior test-time scaling} to standard training, with gains especially pronounced on those instances that can only be solved by either BFS or DFS. In this \countdown task, we observe that ModC with separate models outperforms ModC with prefixes, suggesting that knowledge transfer across modes is less crucial. 

We apply ModC to real-world LLM reasoning datasets with both short- and long-form CoT. Our experiments fine-tune two different base LLMs, Qwen2.5-Base and OLMo2-Base across model scales from 0.5B to 7B, using distillation from two distinct teachers, each representing a different mode. Across all settings, we observe a clear and \textit{consistent} trend: ModC significantly improves test-time scaling compared to both standard mixed-teacher training and the best single-teacher baseline. In particular, on Qwen2.5-7B, ModC training yields up to {4$\times$ efficiency gains} on both long-CoT (Figure~\ref{fig:teaser}) and short-CoT (Figure~\ref{fig:teachers_teacher}). Notably, standard training often fails to exploit teacher diversity, with single-teacher models outperforming those trained on mixtures -- a counterintuitive outcome, since more diverse data should in principle help. ModC, by contrast, effectively harnesses this diversity, translating it into stronger test-time scaling. In other words, diverse training data is most useful when paired with mechanisms that preserve and control its modes.

Finally, we turn to ask: can ModC better harness the implicit diversity of existing post-training datasets (such as NuminaMath; \citealp{numina_math_datasets}) that standard training might be missing? However, applying mode conditioning requires knowing the modes apriori as well as mode annotations on the training data. We find that using a natural idea of gradient clustering (inspired by \citet{Xia2024LESSSI,Jung2025PrismaticSG}) allows us to effectively approximate underlying modes. Once again, we see consistent gains with ModC achieving up to 10\% gains over standard training on the NuminaMath dataset with no additional side information.

Finally, we show that ModC improves reinforcement learning (RL) and can further boost diversity-inducing RL methods. While RL brings both models to the same \passat{1}, ModC excels immediately at $k=2$ with higher \passat{k} scores. This result shows that unlike standard SFT, ModC successfully enriches the solution space without degrading the accuracy of its top output. We see similar benefits of ModC on top of \passat{k} RL \cite{Chen2025PasskTF}, which explicitly prevents diversity collapse during RL. This suggests that interventions designed to prevent diversity collapse during RL can be further boosted by SFT interventions such as our ModC.

In summary, we make the following contributions. 
\vspace{-4pt}
\begin{enumerate}[leftmargin=1.5em]
    \item \textbf{Introducing the mode-conditioning (ModC) framework.} We propose a simple but powerful paradigm to address diversity collapse in LLM reasoning and improve test-time scaling (Section~\ref{sec:modc}). ModC explicitly allocates test-time compute across reasoning modes. We propose two training methods to allow for such test-time allocation: (i) specialist models and (ii) mode-specific prefixes. 
    \item \textbf{ModC demonstrates consistent gains across tasks.} Through controlled graph-search experiments (Section~\ref{sec:countdown}) and large-scale reasoning benchmarks (short- and long-form CoT, distillation from multiple teachers) (Section~\ref{sec:math}), we show that ModC achieves substantial and consistent improvements in test-time scaling, including up to {4$\times$ efficiency gains}. We also carefully analyze tradeoffs between different ModC training methods, effect of model size, data composition etc. 
    \item \textbf{ModC on training data without explicit modes.} We find that gradient clustering provides an effective way to automate mode-conditioning without requiring explicit mode labels, yielding consistent test-time gains. These results suggest that standard training leaves substantial performance untapped by failing to fully exploit diverse data—an inefficiency that the ModC framework directly addresses.
    \item \textbf{The benefit of ModC holds on top of standard RL and diversity-inducing RL.} Unlike standard SFT, ModC successfully enriches the solution space and mitigates diversity collapse during RL. Results on top of \passat{k} RL suggest that interventions designed to prevent diversity collapse during RL can be further boosted by SFT interventions such as our ModC.
\end{enumerate}

\section{The Mode-conditioning framework}
\label{sec:modc}
\subsection{Preliminaries}

Large language models (LLMs) generate outputs by sampling from a probability distribution over continuations. In more complex tasks, instead of producing a single output, we can allocate additional \emph{test-time compute} by drawing $k$ independent samples for the same input and selecting the best candidate. This strategy, known as \emph{parallel scaling}, is especially effective in tasks where solutions can be automatically verified (e.g., mathematics or programming). The performance of this standard approach is captured by the $\text{\passat{k}}_\texttt{std}$ metric on an input $x$: 
\begin{align}
\text{\passat{k}}_\texttt{std}(x) = 1 - (1 - p_x)^k,
\end{align}
where $p_x$ is the \passat{1} or probability of sampling a trajectory that is successful on input $x$. 

In practice the gains with parallel scaling depend strongly on the underlying success probability $p_x$. Modern training pipelines such as supervised fine-tuning and reinforcement learning often induce \emph{mode collapse}, where the model commits to a small set of strategies~\citep{dang2025weightensemblingimprovesreasoning,yue2025doesreinforcementlearningreally,sessa2024bondaligningllmsbestofn, chow2024inferenceawarefinetuningbestofnsampling}. On some prompts, this collapse drives the probability $p_x$ of sampling a successful strategy to be very small, so that an impractically large number of samples is required to obtain good performance. 

\subsection{Mode-conditioned test-time scaling}
One approach is to modify the finetuning objective to prevent collapse and maintain higher $p_x$. We take a complementary route: rather than sampling from a single collapsed distribution, we explicitly allocate test-time compute across \emph{diverse modes}, enforcing coverage so samples include not only the dominant strategy but also alternatives that may succeed where it fails.

Consider two modes with success probabilities $p_{1,x}$ and $p_{2,x}$. If we split the budget evenly, sampling $k/2$ trajectories from each mode, the resulting probability of solving input $x$ is
\begin{align}
\text{\passat{k}}_{\texttt{ModC}}(x) \;=\; 1 - (1-p_{1,x})^{k/2}(1-p_{2, x})^{k/2}.
\end{align}

In contrast, suppose the model does \emph{not} know \textit{a priori} which mode is better for $x$. Let $w_x \in [0,1]$ denote the (random) probability with which the model uses the first mode, and assume that the model is unbiased between the two modes, in the sense that $\mathbb{E}[w_x] = 1/2$ (e.g., $w_x$ is drawn from any distribution centered at $0.5$). A single sample from the model succeeds with probability
$w_x p_{1,x} + (1-w_x) p_{2,x}$,
so we have
\begin{equation}
\text{\passat{k}}_{\texttt{std}}(x; w_x)
= 1 - (1 - w_x p_{1,x} - (1-w_x) p_{2,x})^k.
\end{equation}
The function $w \mapsto 1 - (1-p_{2,x} - (p_{1,x} - p_{2,x}) w)^k$ is concave on $[0, 1]$, so by Jensen's inequality
\begin{equation}
\mathbb{E}_{w_x}\big[\text{\passat{k}}_{\texttt{std}}(x; w_x)\big]
\;\le\;
\text{\passat{k}}_{\texttt{std}}\big(x; \mathbb{E}[w_x]\big)
= \text{\passat{k}}_{\texttt{std}}(x; 1/2).
\end{equation}
When $w_x = 1/2$, the single-sample success rate is $p_x = (p_{1,x} + p_{2,x})/2$. Whenever $p_{1,x} \neq p_{2,x}$, we have
\begin{equation}
(1-p_{1,x})^{k/2}(1-p_{2,x})^{k/2} < (1-p_x)^k,
\end{equation}
which implies
\begin{tcolorbox}[left=7mm,right=0mm,top=0mm, bottom=0mm, colback=blue!5!white,colframe=white]
\begin{equation}
\text{\passat{k}}_{\texttt{ModC}}(x)
>
\text{\passat{k}}_{\texttt{std}}(x; 1/2)
\;\ge\;
\mathbb{E}_{w_x}\big[\text{\passat{k}}_{\texttt{std}}(x; w_x)\big].
\end{equation}
\end{tcolorbox}
In other words, even when the model's mode-usage follows \emph{any} distribution centered at $0.5$, explicitly allocating compute evenly across modes is strictly better than sampling from the model's own uncertain mixture.

But how do we explicitly sample from different modes in practice? A naive baseline is to prompt the model with instructions to use different strategies. However, this approach is unreliable: it is unclear how to phrase the prompts, and the model may not consistently follow them.  

\subsection{Mode-conditioned training}
Instead of relying on ad-hoc prompting to elicit different behaviors, we consider scalable training objectives that explicitly enforce control over modes. This provides a reliable lever at test time for allocating compute across diverse strategies. We explore two natural instantiations: training with separate specialist models and training with prefixes within a single model.  

In this section, we assume that the relevant modes are known \textit{a priori} and that training data can be annotated with the mode used to generate each trajectory. While this assumption is convenient for exposition and testing the benefits of our paradigm, it is not strictly necessary. In practice, one could imagine automated approaches for mode discovery and annotation, for example by clustering trajectories using gradient-based similarity measures or other unsupervised techniques. We return to this point in \S\ref{sec:automated} where we discuss how mode-conditioned training can be extended to settings where the modes are not explicitly labeled in the data.

\textbf{Mode-conditioned training with separate models. \ \ }  
The most direct approach is to train distinct models, each specialized to a particular mode of reasoning. Concretely, the training data is partitioned into subsets corresponding to different strategies, and a separate model is trained on each subset while keeping total training data and compute constant. At test time, the sampling budget is divided across the specialists (e.g., $k/2$ samples from each in the two-mode case). This design ensures strong specialization and reduces correlated errors, which translates into more effective parallel scaling.  

\textbf{Mode-conditioned training with prefixes. \ \ }  
While separate models improve diversity, they prevent knowledge sharing across modes. This is a significant drawback in realistic reasoning tasks, where different strategies often rely on common linguistic or mathematical foundations. 

To overcome this, we draw inspiration from the literature on steering model behavior via \textit{explicit conditioning tokens}, a technique used widely in controlled text generation~\citep{keskar2019ctrlconditionaltransformerlanguage} and instruction tuning. We prepend discrete condition tokens (e.g., \texttt{[Mode 1]}, \texttt{[Mode 2]}) to the input, training the model to associate each prefix with a distinct reasoning strategy. At inference, balanced compute allocation is enforced by sampling evenly across the conditioning prefixes. This allows the model to specialize into distinct modes while still sharing knowledge across them, making it more scalable than training separate specialist models.

\section{Investigating mode coverage in parallel sampling}
\label{sec:countdown}

\subsection{The \countdown task}

\countdown is a generalization of \textit{Game of 24}, where a model must find a sequence of arithmetic operations to transform a set of starting numbers into a target value~\citep{Gandhi2024StreamOS}. Given several starting numbers, the model can apply operations $\{+, -, \times, \div\}$ to reach a target. For example, given $\{10, 10, 4, 6\}$ with target $16$, one solution is $(10 \times 10 - 4) \div 6 = 16$.

This task naturally admits two distinct problem-solving modes: depth-first search (DFS) and breadth-first search (BFS). This allows us to precisely control and examine which mode is used. Solutions are easily verifiable by checking if operations reach the target, which makes it an ideal testbed for studying test-time scaling with parallel sampling.

\subsection{Why mode coverage is important?}

In principle, both BFS and DFS are complete search algorithms capable of finding any solution, so why would mode coverage matter for this task? For real-world problems, however, computational constraints require using heuristics to make the search tractable. 
Following \citet{Gandhi2024StreamOS}, we use heuristics to guide and prune the search space, which introduces an important asymmetry: with heuristic pruning and search budget constraints, each algorithm excels on different problem instances -- some problems become solvable only by the oracle DFS while others only by the oracle BFS. Since we cannot know \textit{a priori} which algorithm will succeed for a given problem, maintaining coverage of both modes during test-time sampling becomes crucial for achieving high success rates. 

To evaluate test-time scaling, we report Pass@$k$ metrics on two held-out test sets. We first create a \textbf{\textit{natural}} test set of 500 problems by randomly selecting unseen target numbers and valid starting numbers that can reach the target. Second, we subsample an \textbf{\textit{adversarial}} test set designed to require mode diversity: we filter for problems where either oracle BFS or oracle DFS (but not both) achieves less than $5\%$ success rate across multiple runs. This adversarial set directly tests mode coverage -- each problem is effectively solvable by only one algorithm, so achieving high accuracy requires sampling from both DFS and BFS modes when we do not know \textit{a priori} which one works better. We expect that the benefit from explicit test-time balancing is much larger on the adversarial test set.

\subsection{Mode coverage}

We use rejection sampling to create our training set, keeping only instances where at least one search algorithm (oracle DFS or BFS) successfully finds a solution. Specifically, we uniformly generate a target number from 1 to 200 and four starting numbers that can reach the target, and uniformly choose DFS and BFS and one of the search heuristics for guidance and pruning. We note that DFS has a higher overall success rate, so our final training set ends up consisting of 163K problems, with 97K DFS solutions and 65K BFS solutions. Each training example includes the input, the search trajectory, and final operations. We train Qwen2.5-Base models (0.5B--7B) for 4 epochs.

\begin{figure*}[!th]
\centering
    \includegraphics[width=\linewidth]{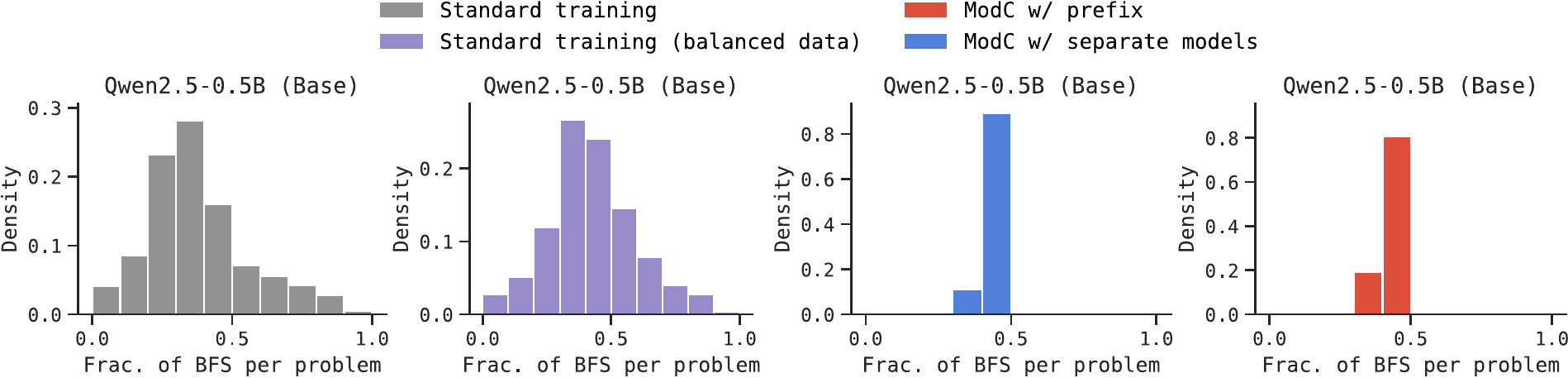}
\caption{\label{fig:bfs_dfs} \textbf{Standard training fails to balance diverse modes \textit{per problem} under repeated sampling.} This issue does not go away with balanced training data. Instead, ModC explicitly targets and successfully achieves balanced test-time compute allocation. }
\end{figure*}

We first examine whether models trained on mixed DFS and BFS data can learn to balance the two algorithms under repeated sampling. Figure~\ref{fig:bfs_dfs} shows the fraction of BFS used by the model for each test problem. We observe that standard training on the mixture of both algorithms (shown in gray) tends to bias toward one algorithm for many test problems, i.e., either predominantly using DFS (low BFS fraction) or BFS (high BFS fraction), rather than balancing both.

\textbf{Effect of diversity of training data. \ \ } Recall that we use rejection sampling to create our training set, which naturally biases the training data towards the on-average more promising algorithm (i.e., DFS in this case). What if we had 50-50 data for DFS and BFS? To answer this, in Figure~\ref{fig:bfs_dfs} we also plot the distribution for this standard training with balanced data. We see that the distribution is less skewed, but still many problems have extremely imbalanced allocation of test-time compute.

\textbf{ModC balances test-time compute allocation. \ \ } In contrast, ModC with explicit balanced allocation achieves the desired behavior. We see that for both training separate models (shown in blue) and training with prefixes (shown in red), the fraction of BFS per problem is concentrated around 0.5, which demonstrate nearly perfect balance.

\begin{figure*}[!th]
\centering
\begin{subfigure}[b]{\textwidth}
    \centering
    \includegraphics[width=\linewidth]{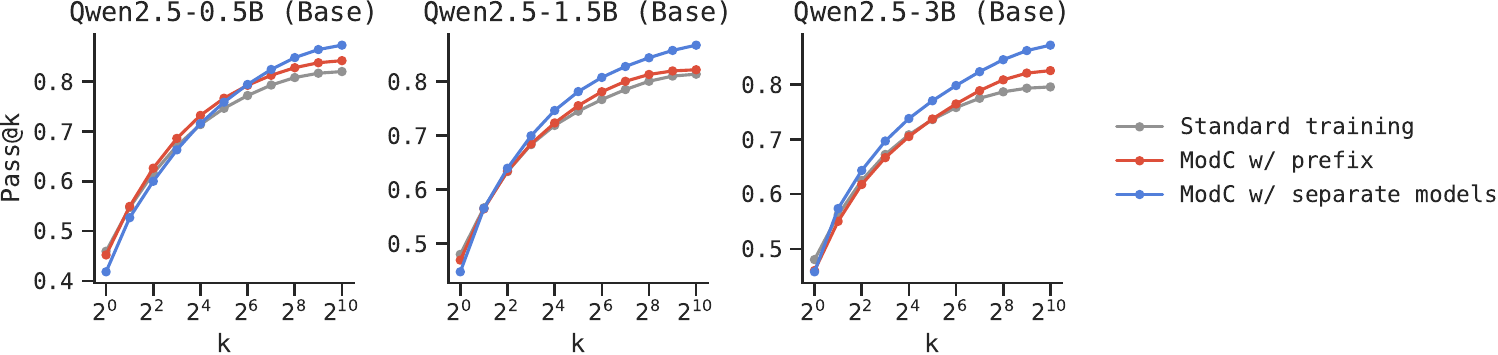}
    \caption{Pass@$k$ performance on \countdown natural test set}
\end{subfigure}
\hfill \vspace{0pt}
\begin{subfigure}[b]{\textwidth}
    \centering
    \includegraphics[width=\linewidth]{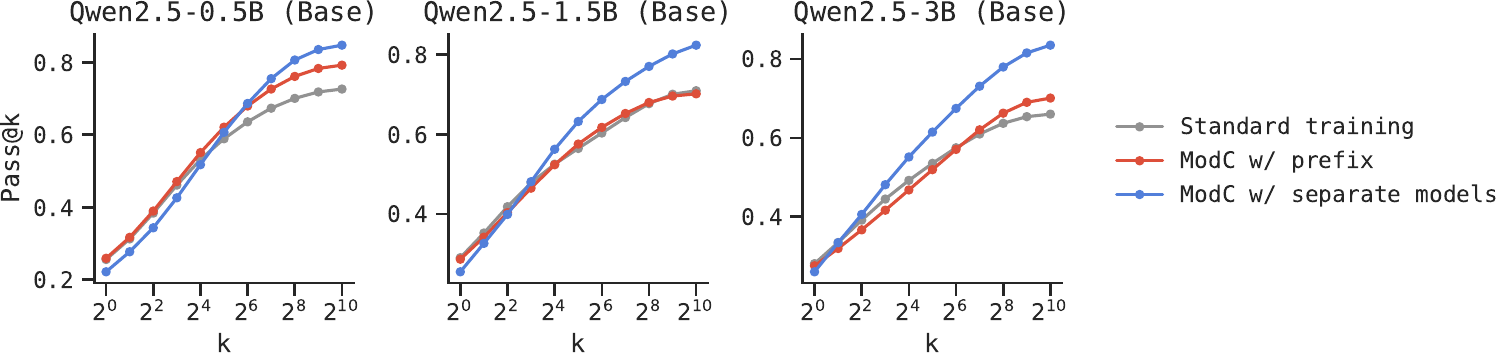}
    \caption{Pass@$k$ performance on \countdown adversarial test set}
\end{subfigure}
\caption{\textbf{Balanced test-time allocation improves scaling.} (a) On the natural test set of \countdown, balanced test-time allocation with ModC shows consistent improvements as $k$ increases. (b) On the adversarial test set where each problem is challenging for one one algorithm (oracle DFS or BFS), the gains from enforced mode diversity are even more pronounced.}
\label{fig:countdown_passk}
\end{figure*}

\subsection{From mode coverage to \passat{k}}

In this part, we show that ModC dramatically improves test-time scaling, particularly on problems that require diverse algorithmic modes. We start with ModC with separate models. Figure~\ref{fig:countdown_passk} shows the results on the natural and adversarial test set of \countdown across model scales. While \passat{1} is comparable or slightly lower for separate models, the scaling behavior dramatically improves by up to $8\%$ for \passat{1024}. This gap is consistent across all model sizes, with separate models maintaining superior scaling curves. 
The advantages are even more pronounced on the adversarial test set, where we see that the gap boosts up to $20\%$ for \passat{1024}. This shows that mode balance is indeed crucial for test-time scaling -- problems that DFS finds hard to solve are potentially solvable for BFS, which enable better coverage of the problem space.
On the other hand, Figure~\ref{fig:countdown_passk} shows that ModC with prefix with balanced allocation outperforms standard training for most scales. As a control, we try random partitioning the training data into two groups, which sometimes shows gains but does not outperform ModC (see details in \S\ref{subsec:countdown-additional-results}). 

\begin{figure*}[t!]
\centering
\begin{subfigure}[b]{\textwidth}
    \centering
    \includegraphics[width=0.86\linewidth]{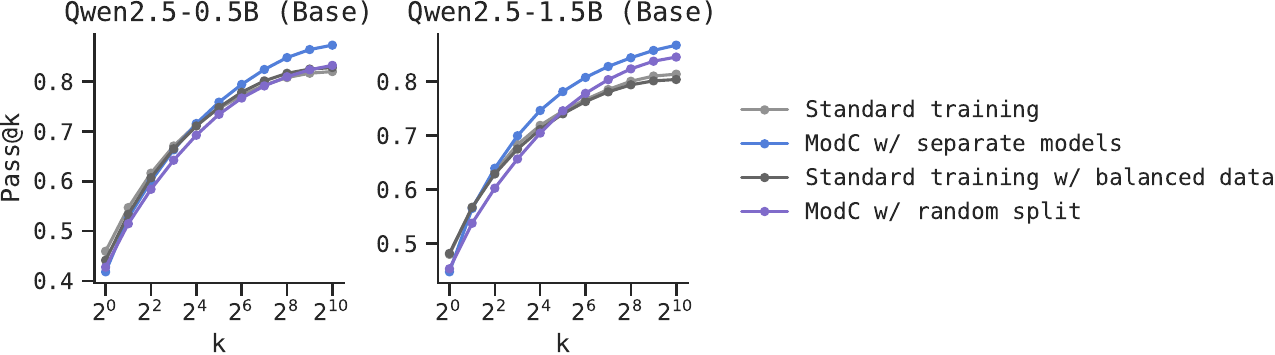}
    \caption{Pass@$k$ performance on \countdown natural test set}
\end{subfigure}
\hfill \vspace{0pt}
\begin{subfigure}[b]{\textwidth}
    \centering
    \includegraphics[width=0.86\linewidth]{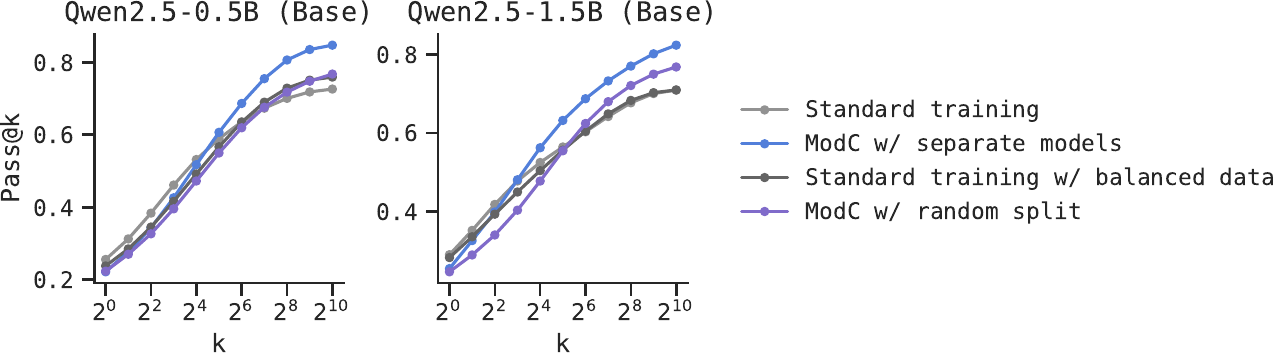}
    \caption{Pass@$k$ performance on \countdown adversarial test set}
\end{subfigure}
\caption{\textbf{Ablation studies on \countdown.} ModC with random paritioning sometimes shows gains but does not outperform ModC with DFS/BFS partition. Balanced training data DFS/BFS distribution does not show gains compared to standard training. }
\label{fig:countdown_passk_ablation}
\end{figure*}

\subsection{Additional ablations for \countdown}
\label{subsec:countdown-additional-results}

We see that ModC with DFS/BFS partition improves test-time scaling on \countdown. As a control, we also try random partitioning the training data into the same number of groups. From Figure~\ref{fig:countdown_passk_ablation}, we see that ModC with random paritioning sometimes shows gains but does not outperform ModC with DFS/BFS partition. Another baseline we try is to enforce 50-50 distribution of DFS and BFS in the training data, which we do not see any gains compared to standard training.

\section{Mode-conditioning improves math post-training}
\label{sec:math}
We saw how ModC improves \passat{k} performance on \countdown and is superior for parallel scaling. In this section, we evaluate ModC when post-training for math reasoning. We investigate both long and short CoT reasoning across multiple model families and scales.

\subsection{Distillation from multiple teachers}
We start with a natural source of diverse modes: distillation from multiple stronger teacher models. This setting is particularly relevant given the existence of strong models with distinct reasoning and response styles. Recent work typically selects the single teacher that provides the best distillation performance \cite{Muennighoff2025s1ST,Guha2025OpenThoughtsDR}. 
We test whether explicitly balancing compute across multiple teacher strategies improves performance. 
In each setting, we collect CoT reasoning traces from two teacher models for mathematical problems, where each teacher demonstrates distinct problem-solving styles. Following the methods established in Section~\ref{sec:countdown}, we compare four training strategies: (1) \textit{mixed teachers}: mixing all teacher data together, (2) \textit{best teacher}: training on the teacher data that produce the best performance (3) \textit{ModC with separate models}: training separate models on each teacher's data independently, and (3) \textit{ModC with prefixes}: prepending teacher identity tokens (e.g., \texttt{using [teacher]-style reasoning}) to each CoT and using balanced test-time allocation where we sample equally from each teacher condition.

\begin{figure*}[t!]
    \includegraphics[width=\linewidth]{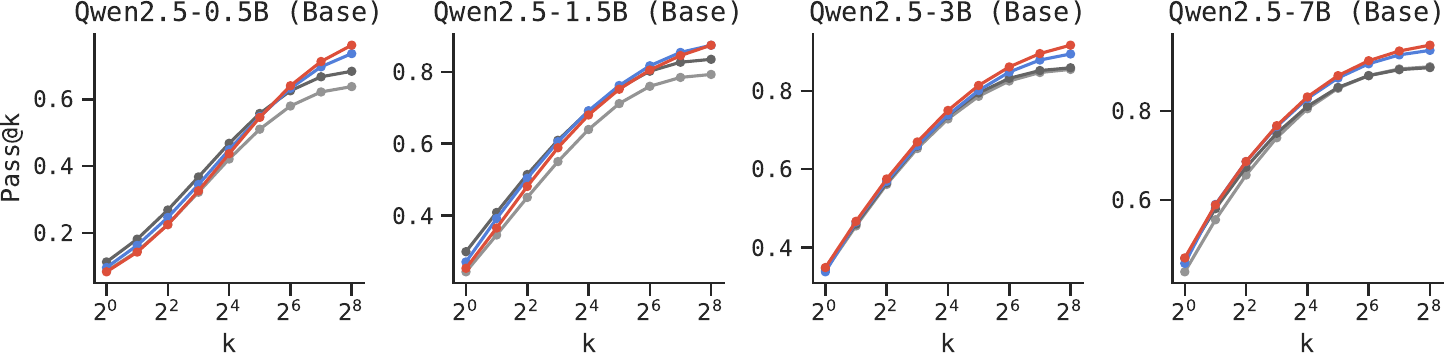}
    \includegraphics[width=0.9\linewidth]{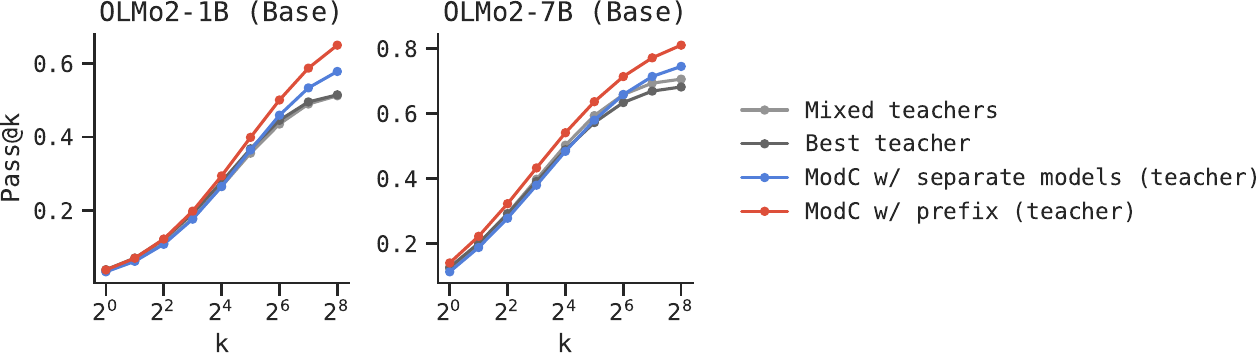}
\caption{\label{fig:teachers_teacher} \textbf{ModC improves short CoT reasoning.} \passat{k} on MATH500. Naively mixing teacher data underperforms the single-teacher baseline, while ModC shows consistent gains. ModC with prefixes generally works better than ModC with separate models underscoring the importance of sharing knowledge across modes (teacher strategies) in math reasoning.}
\end{figure*}

\subsection{Short chain-of-thoughts}
\label{subsec:short-cot-teachers}

\textbf{Experimental setup. \ \ } We first experiment with post-training NuminaMath \cite{numina_math_datasets} dataset where the chain-of-thought completions are relatively short and do no involve long thinking.
We use the SFT traces distilled from two teacher models: DeepSeek-R1 \cite{DeepSeekAI2025DeepSeekR1IR} and GPT-OSS-120B \cite{openai2025gptoss120bgptoss20bmodel}, with the problems from NuminaMath. 
For evaluation, we use MATH500 \cite{hendrycksmath2021} and measure \passat{k}. We post-train Qwen2.5-Base (0.5B--7B) and OLMo2-Base (1B--7B) models for 4 epochs. We tune the learning rate $\in\{1e\text{-}4, 1e\text{-}5\}$ and use AdamW optimizer with global batch size of 256. 

\textbf{Naive data mixture underperforms. \ \ }Figure~\ref{fig:teachers_teacher} compares the \passat{k} curves on MATH500 for the four training strategies mentioned above. We observe that naively mixing data from both teachers either underperforms or is at best comparable to the stronger single-teacher baseline. This is consistent with prior intuition that training on the best teacher can be more effective than mixing teachers.

\textbf{ModC training unlocks superior test-time scaling. \ \ } On the other hand, training on both teachers' data but with mode conditioned (ModC) training and inference unlocks better test-time scaling (Figure~\ref{fig:teachers_teacher}). The gains are consistent across model families (Qwen and OLMo2) and scales (0.5B to 7B), offering up to 10\% gain on Qwen2-0.5B and 15\% on OLMo-2-7B.
Comparing the two variants of ModC, we see that ModC with prefixes generally outperforms ModC with separate models suggesting that knowledge sharing across modes is crucial for math tasks.

\subsection{Long chain-of-thoughts}

We now examine long CoT reasoning, where models usually spend tens of thousands of tokens on extended reasoning before producing the answer on more challenging problems. 

\begin{wrapfigure}[21]{r}{0.47\textwidth}
\vspace{-0pt}
\centering
\includegraphics[width=0.6\linewidth]{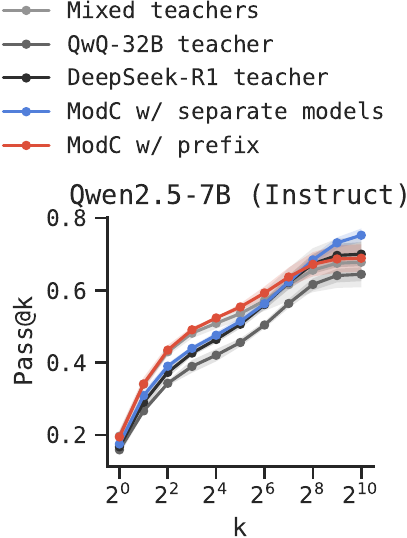}
\caption{\label{fig:teachers_teacher_openthoughts} \textbf{ModC improves long CoT reasoning.} \passat{k} on AIME 2025. Standard training with multiple teachers fails to outperform the best single teacher, but ModC with multiple teachers surpasses single teacher.}
\label{fig:defenses}
\end{wrapfigure}

\textbf{Experimental setup. \ \ } We use the subset of problems from OpenThoughts-3 \cite{Guha2025OpenThoughtsDR} that they did ablation studies for the teacher models with. Specifically, solutions are from two teachers: QwQ-32B \cite{qwq32b} and DeepSeek-R1 \cite{DeepSeekAI2025DeepSeekR1IR}. For evaluation, we use AIME 2025 and measure \passat{k}. Following \citet{Guha2025OpenThoughtsDR}, we initialize model weights with Qwen2.5-7B-Instruct.

\textbf{4$\times$ efficiency gains with ModC. \ \ }
Figure~\ref{fig:teachers_teacher_openthoughts} shows similar patterns to short CoT: ModC achieves consistently higher \passat{k} than both single-teacher and mixed-teacher baselines. Even in the long CoT setting with extremely large token budgets per sample, standard training fails to adequately cover multiple modes—mixed training again does not outperform the best single teacher. In contrast, ModC not only provides an effective way to learn from multiple teachers and surpass each individual teacher, but also delivers substantial efficiency gains: it matches the Pass@1024 of standard training with only $k=256$ samples, yielding nearly \textbf{4$\times$ faster inference}, while simultaneously pushing the maximum achievable Pass@$k$. We 

\section{Automated mode discovery}
\label{sec:automated}
We have seen that ModC improves test-time scaling across a wide variety of real-world settings. However, previous sections assume access to natural modes in the data: search algorithms in \countdown or teacher identities in multi-teacher distillation. In practice, most real-world training data lacks such clear segregation. Can we extend ModC to work on training data that contains mixed modes but lacks explicit labels? Furthermore, can we discover meaningful modes in training data without apriori knowledge of these modes? We explore both questions in this section, applying gradient clustering to discover and annotate modes in training data.

\subsection{Gradient clustering}
Gradient similarity has been shown effective in understanding training dynamics \cite{Jacot2018NeuralTK}, identifying influential training data \cite{Koh2017UnderstandingBP}, and diversifying data selection \cite{Jung2025PrismaticSG}. That inspires us to test whether gradient cluster can discover meaningful modes in training data that we should condition on. 

For each training example $(x, y)$, we compute gradients with respect to model parameters $$g_\theta(x, y) = \nabla_{\theta} \log p_{\theta}(y|x).$$ To reduce dimensionality, we follow prior works \cite{Xia2024LESSSI,Jung2025PrismaticSG} to apply Rademacher random projection \cite{Park2023TRAKAM} to each gradient vector. Once we get the projected gradient vectors for all training samples, we cluster the vectors into $C$ clusters, and all samples in the same cluster belong to one ``mode'' based on which we apply ModC.

\subsection{Gradient clustering recovers teacher identity}
\begin{figure*}[!th]
\centering
    \includegraphics[width=\linewidth]{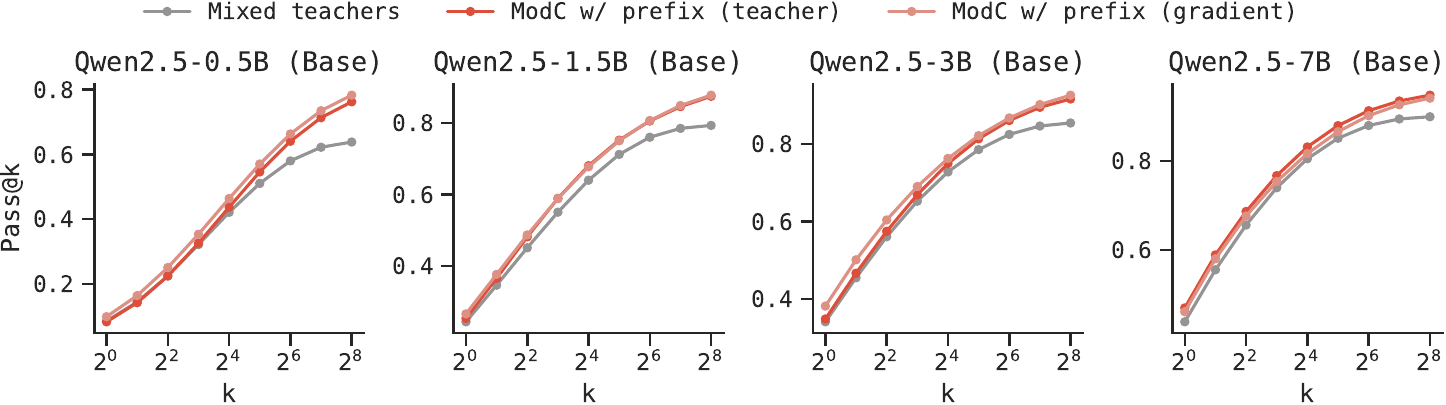}
\caption{\label{fig:teachers_cluster} \textbf{Validating gradient clustering on multi-teacher data.} ModC on training data that is distilled from multiple teachers outperforms standard training even without access to teacher identity annotation on training data. ModC with gradient clustering almost completely matches ModC with access to teacher annotations.}
\end{figure*}

We first validate gradient clustering on multi-teacher data where ground-truth labels exist. Using the short CoT dataset from Section~\ref{subsec:short-cot-teachers}, we compute gradients using Qwen2.5-Base 1.5B and apply clustering. We first observe that gradient clustering achieves $98.7\%$ F1 score in recovering teacher assignments. More importantly, Figure~\ref{fig:teachers_cluster} shows that ModC with these automatedly discovered gradient-based clusters yields nearly identical test-time scaling benefits as using true teacher labels. This confirms that gradient patterns effectively capture the underlying modes.

\subsection{Gradient clustering improves post-training on general data}
\begin{figure*}[!th]
\centering
    \includegraphics[width=\linewidth]{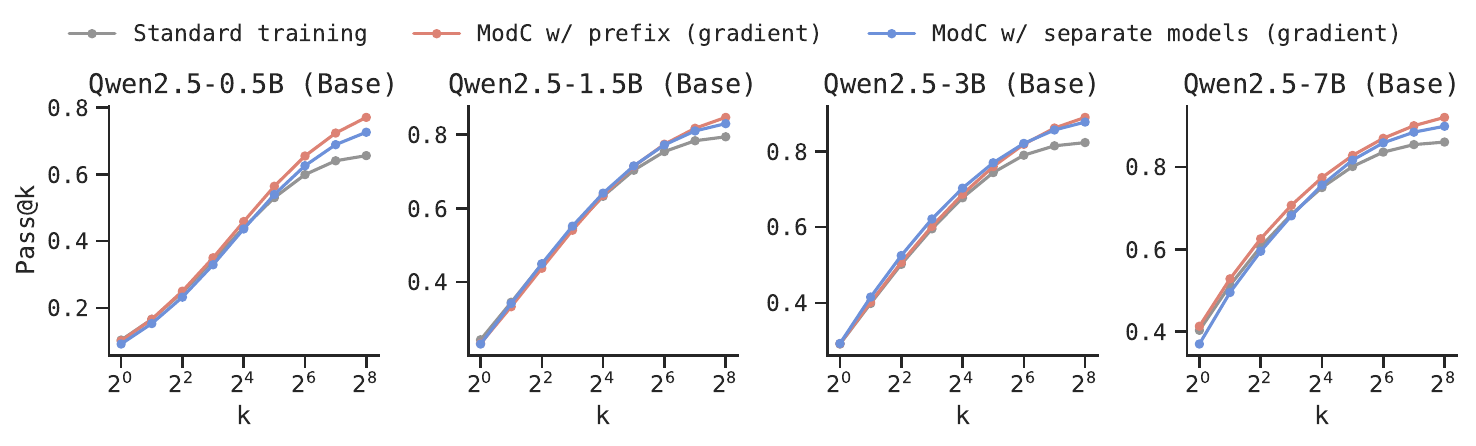}
\caption{\label{fig:numina_cluster} \textbf{ModC on automatedly discovered modes via gradient clustering improves short CoT.} \passat{k} on MATH500 for Qwen2.5-Base models finetuned with standard finetuning and ModC.}
\end{figure*}

We apply gradient clustering to NuminaMath, a real-world dataset that is probably quite diverse but we lack a clear sense of what modes exist. Surprisingly, we see that training with ModC on automatedly discovered modes (via gradient clustering) yields significant improvements. Figure~\ref{fig:numina_cluster} shows that ModC consistenty improves \passat{k} compared to standard training across model scales.

\section{Mode-Conditioning Improves Reinforcement Learning}

Finally, we turn to verify if the benefit of ModC holds after RL, where the distribution becomes sharper and diversity decreases. For RL methods: we consider (1) standard RL, where the distribution becomes sharper and diversity decreases, and (2) \passat{k} RL \cite{Chen2025PasskTF}, which can improve both \passat{1} and \passat{k}. For each RL method, we initialize the policy from either the standard SFT models or ModC models. For the prefix variant, we believe we need prefix-following rewards to make sure that the learned prefix-mode binding is maintained. Therefore, we focus on the separate models variant given its simplicity and leave RL on the prefix variant for future work.

\begin{wrapfigure}[13]{r}{0.5\textwidth}
\vspace{-10pt}
\centering
\includegraphics[width=\linewidth]{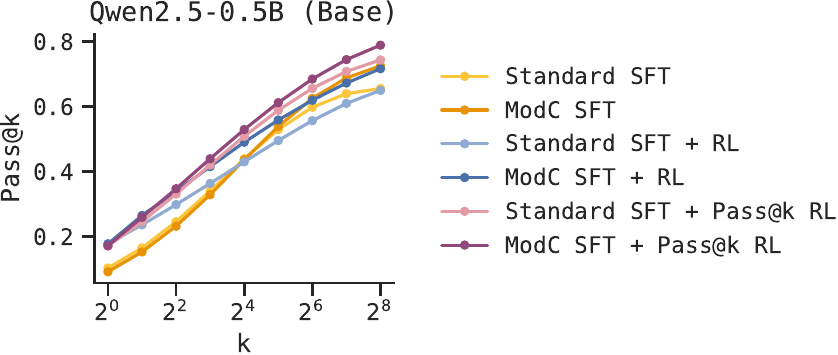}
\caption{\label{fig:rl} \textbf{ModC improves long CoT reasoning.} \passat{k} on MATH500. While RL brings both models to the same \passat{1}, ModC excels immediately at $k=2$ with higher \passat{k} scores. }
\label{fig:defenses}
\end{wrapfigure}

Figure~\ref{fig:rl} shows the results of ModC vs standard SFT followed by different RL methods. Crucially, while RL brings both models to the same \passat{1}, ModC excels immediately at $k=2$ with higher \passat{k} scores. This result shows that unlike standard SFT, ModC successfully enriches the solution space without degrading the accuracy of its top output. Moreover, we see that this observation also holds for \passat{k} RL, which explicitly prevents diversity collopse during RL training. This suggests that interventions designed to prevent diversity collapse during RL can be further boosted by SFT interventions such as our ModC.

\section{Related work}

\textbf{Improving parallel test-time scaling. \ \ } Repeated sampling significantly improves the performance of LLMs especially in domains like reasoning and coding~\citep{wang2023selfconsistencyimproveschainthought,brown2024largelanguagemonkeysscaling,rozière2024codellamaopenfoundation}. To decide the final answer from the $k$ attempts, one can use either majority voting~\citep{wang2023selfconsistencyimproveschainthought} or a verifier (especially in the code and scientific discovery domains)~\citep{wang2024mathshepherdverifyreinforcellms, alphaevolve}.
However, a series of works~\citep{cobbe2021trainingverifierssolvemath,dang2025weightensemblingimprovesreasoning} have identified issues during post-training of LLMs that impair the diversity in model generation, consequently affecting the efficacy of test-time scaling. ~\citet{huang2024selfimprovementlanguagemodelssharpening} attribute this to the sharpening effect whereas \citet{chu2025sftmemorizesrlgeneralizes} highlight memorization as a root cause. 

Lot of recent works have in turn have proposed fixes for improved parallel test-time scaling. \citet{beeching2024scalingtesttimecompute,snell2024scalingllmtesttimecompute} propose modifications to beam search to explciitly optimize for diversity amongst the candidates. \citet{Wang2025DiversifiedSI,Hughes2024BestofNJ} propose diverse prompting to improve test-time scaling.
Taking a step back, \citet{sessa2024bondaligningllmsbestofn, chow2024inferenceawarefinetuningbestofnsampling,chen2025rethinkingfinetuningscalingtesttime} explicitly optimize for best-of-k performance during the finetuning process. ~\citet{dang2025weightensemblingimprovesreasoning} propose a simple fix of ensembling the finetuned weights with the base model to mitigate diversity collapse. \citet{goyal2025distilledpretrainingmodernlens} further take a  step back and propose pretraining with logit distillation to improve parallel test-time scaling behaviors. 
In contrast, in this work we propose a more data centric conditioning of the finetuning process to explcitly encode specialist modes in the model. 

\textbf{Creativity and diversity of language models. \ \ } Recent work has evaluated creativity and diversity in language models with mixed findings. Models underperform human experts in creative writing \citep{chakrabarty24artifice} and humor generation \citep{zhang24humor}. While \citet{anderson24homogenization} find that language models increase individual idea divergence but causes group-level homogenization, \citet{si24can} report that LLMs can generate novel research ideas despite feasibility limitations. \citet{Nagarajan2025RollTD} demonstrate that global planning and seed-conditioning are crucial for creative generation. Temperature sampling shows weak correlation with creativity and often introduces incoherence \citep{chen23temperature}.
On the evaluation front, various benchmarks have been proposed to measure output diversity and creativity \cite{mclaughlin2025aidanbench,Zhang2025NoveltyBenchEL,Jansen2024DISCOVERYWORLDAV}. 

\textbf{Specialization in model training. \ \ } In this work, we proposed \emph{ModC} where a single model is explicitly conditioned to learn separate modes of finetuning data. We do this by prepending conditioning tokens (e.g., DFS or BFS) to the reasoning traces. Closest to our work is Mixture-of-Experts~\citep{shazeer2017outrageouslylargeneuralnetworks,jiang2024mixtralexperts,jelassi2025mixtureparrotsexpertsimprove} where different datapoints are routed to a specific subpart of the model which is expert for the domain of the particular datapoint. However, in contrast in ModC all the datapoints are processed by the whole model and not a subpart. Mixture-of-experts are aimed at reducing the active parameter footprint of the model.

\section{Conclusions and future work}
In this work, we demonstrated that deliberate mode-conditioning (ModC) during training and inference is crucial for unlocking the full benefits of test-time compute scaling. Across both controlled search problems and large-scale reasoning benchmarks, we showed that standard training tends to collapse onto a single strategy, while specialization through mode-conditioning consistently yields superior scaling and more reliable gains. Beyond explicit labels such as teacher identity, we further showed that gradient clustering can automatedly uncover meaningful specializations, making ModC broadly applicable. We finally show that ModC mitigates diversity collapse during both standard RL and diversity-inducing RL. It suggests that interventions designed to prevent diversity collapse during RL can be further boosted by SFT interventions such as our ModC. Looking ahead, exciting directions include extending ModC to a larger number of modes and to richer behavioral dimensions such as reasoning depth or planning style, as well as integrating ModC with adaptive allocation policies that learn how to optimally divide compute across modes at test time. Another promising avenue is further improving ModC with reinforcement learning, where balanced modes could encourage diverse exploration early in training and be gradually relaxed. Together, these directions position mode-conditioning as a general and effective principle for building more reliable and powerful reasoning systems.



\section*{Acknowledgments} 
We wish to thank Vaishnavh Nagarajan, Ziqian Zhong, and Jacob Springer for their discussion and feedback on the draft. We gratefully acknowledge support from Schmidt Sciences, NSF, Apple, Open Philanthropy, Google. We would like to thank CMU FLAME Cluster for providing generous compute. 

\bibliography{iclr2026_conference}
\bibliographystyle{iclr2026_conference}
\clearpage
\appendix


\begin{figure*}[ht!]
    \includegraphics[width=\linewidth]{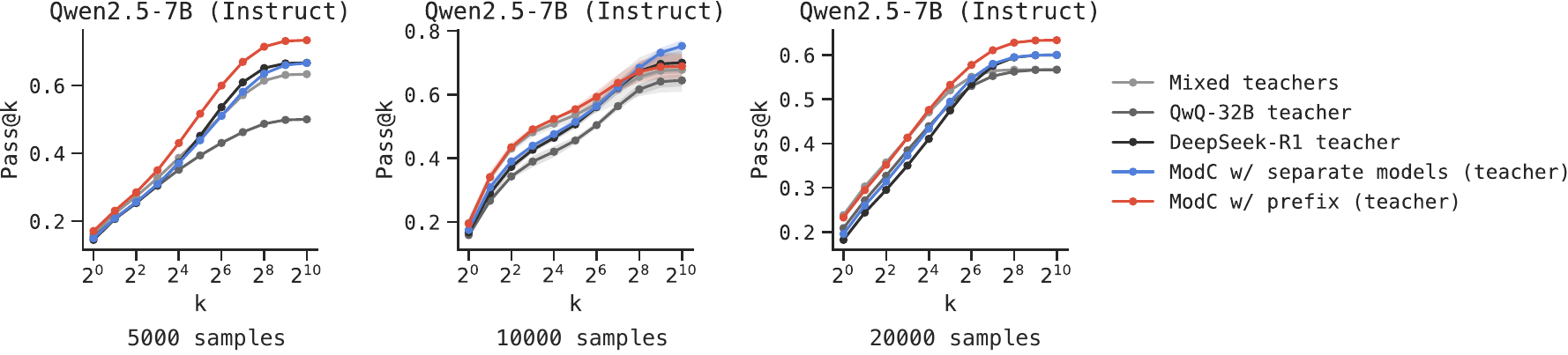}
\caption{\label{fig:data-scales-openthoughts} \textbf{Effect of data scale.} \passat{k} on AIME 2025. ModC demonstrates consistent benefits across all scales, with the largest gains observed at the smallest data scale. }
\end{figure*}

\section{Additional results on data scales}

We conducted experiments on OpenThoughts-3 across varying data scales, from 5k to 20k samples. ModC demonstrates consistent benefits across all scales, with the largest gains observed at the smallest data scale. 

\end{document}